\documentclass[letterpaper, 10 pt, conference]{ieeeconf}  

\usepackage{booktabs}
\usepackage{multicol}
\usepackage{multirow}
\usepackage{placeins}
\usepackage{graphicx}
\usepackage{amsmath}
\usepackage{amsfonts}
\usepackage{cite}
\usepackage{hyperref}
\usepackage{color}

\definecolor{ForestGreen}{RGB}{34,139,0}
\definecolor{SkyBlue}{RGB}{0,191,255}
\definecolor{Gold}{RGB}{255,215,0}
\usepackage{subfigure}
\usepackage{threeparttable}

\IEEEoverridecommandlockouts                              

\title{\LARGE \bf
Imitation Learning of Hierarchical Driving Model: from Continuous Intention to Continuous Trajectory
}

\author{Yunkai Wang, Dongkun Zhang, Jingke Wang, Zexi Chen, Yue Wang$^{\dagger}$, Rong Xiong
\thanks{Yunkai Wang, Dongkun Zhang, Jingke Wang, Zexi Chen, Yue Wang, Rong Xiong are with the State Key Laboratory of Industrial Control Technology and Institute of Cyber-Systems and Control, Zhejiang University, Hangzhou, China}
\thanks{$^\dagger$ Corresponding author, {\tt\small wangyue@iipc.zju.edu.cn}
}}

\begin{document}

\maketitle
\thispagestyle{empty}
\pagestyle{empty}

\begin{abstract}
One of the challenges to reduce the gap between the machine and the human level driving is how to endow the system with the learning capacity to deal with the coupled complexity of environments, intentions, and dynamics. In this paper, we propose a hierarchical driving model with explicit model of continuous intention and continuous dynamics, which decouples the complexity in the observation-to-action reasoning in the human driving data. Specifically, the continuous intention module takes the route planning map obtained by GPS and IMU, perception from a RGB camera and LiDAR as input to generate a potential map encoded with obstacles and intentions being expressed as grid based potentials. Then, the potential map is regarded as a condition, together with the current dynamics, to generate a continuous trajectory as output by a continuous function approximator network, whose derivatives can be used for supervision without additional parameters. Finally, we validate our method on both datasets and simulator, demonstrating that our method has higher prediction accuracy of displacement and velocity and generates smoother trajectories. The method is also deployed on the real vehicle with loop latency, validating its effectiveness. To the best of our knowledge, this is the first work to produce the driving trajectory using a continuous function approximator network. Our code is available at \url{https://github.com/ZJU-Robotics-Lab/CICT}.
\end{abstract}

\section{Introduction}
Autonomous driving is an appealing research topic in recent years because of its ability to reduce labor costs and traffic accidents. In typical autonomous driving systems, vehicles need to perform observation-to-action reasoning to generate safe and efficient driving in various scenarios. One of the challenges to reduce the gap between the machine and the human level driving is how to endow the system with the learning capacity to deal with the coupled complexity of environments, intentions, and dynamics \cite{kuutti2020survey}.  

Toward this goal, there are several learning paradigms. The most common paradigm is to decompose the observation-to-action reasoning into separated modules e.g., detection, tracking, and planning, which has good generalization given the carefully human-designed system structure, but it calls for intensive human supervision for each module \cite{ding2019safe}. To relieve the difficulty, another paradigm is to model the observation-to-action reasoning as a black box, thus massive labeled data can be automatically generated by simply recording the human driving process \cite{bojarski2016end}. However, this paradigm has much lower generalization capability, partly due to the almost missing of the human system design, losing the human knowledge intrinsically reserved in the system structure \cite{chen2015deepdriving}. 

\begin{figure}[t]
	\centering
	\includegraphics[width=0.49\textwidth]{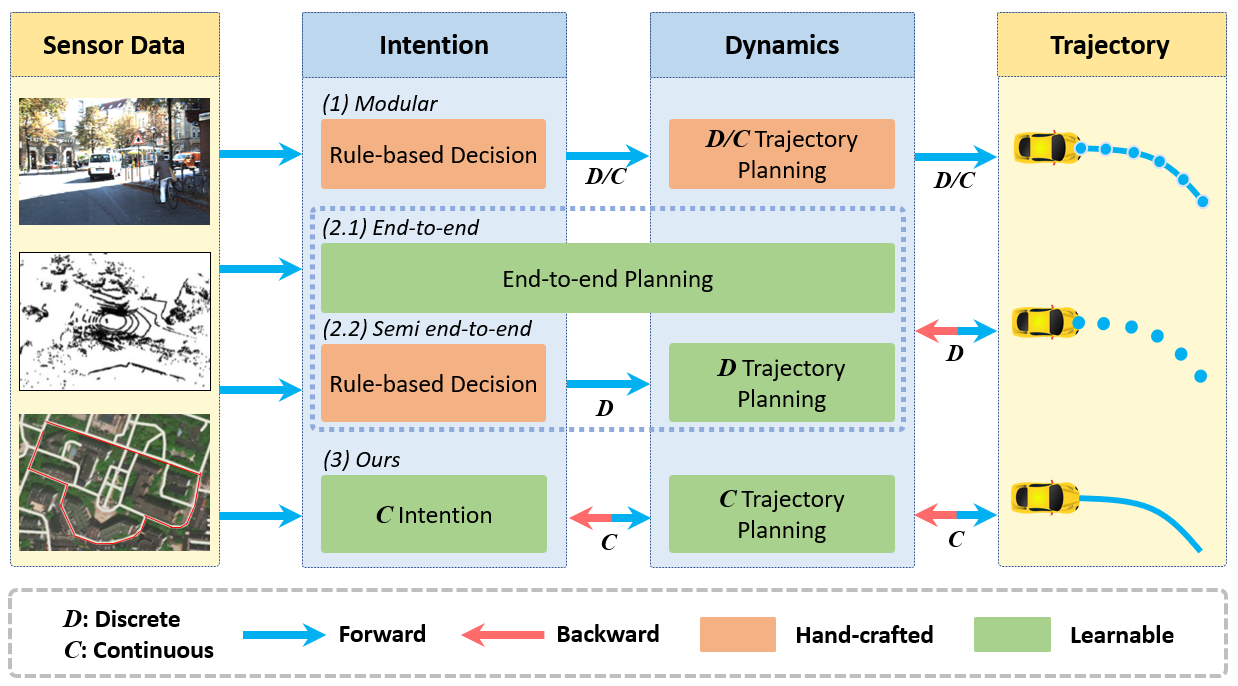}
	\vspace{-0.5cm}
	\caption{Different trajectory planning approaches for autonomous vehicles. (1) Classical modular design; (2.1) End-to-end planning method; (2.2) Simi end-to-end planning method; (3) Our hierarchical model.}
	\label{overview}
\end{figure}

In this paper, we follow the second paradigm to only use the hand-free annotation data to learn a vehicle driving model. At the same time, motivated by the idea in the first paradigm, we also inject human knowledge via the system structure design, which is to explicitly model a hierarchical structure to decouple the intentions and dynamics from the human driving data, as shown in Fig. \ref{overview}. 

Specifically, to model the trajectory, we regard the network as a continuous function approximator, whose high-order derivatives can be obtained analytically without adding model parameters, and they can also be used for supervision to achieve better results. A trajectory can be supervised at any sampling interval without adding model parameters, making it sample efficient. In contrast, the previous works mainly use only the discrete positions as supervision, which loses the continuous motion characteristics of the demonstration data \cite{cai2020vtgnet}. Some previous works also use additional outputs to capture the velocity, but they fail to guarantee the relation of derivatives between position and velocity \cite{cai2020vtgnet} \cite{zhao2019lates}.

For the high-level intention, we follow our previous work \cite{ma2020deepgoal} to model the driving intention generation process as a mapping from coarse route planning to a potential map without time parametrization. Therefore, the intention generation can be regarded as a continuous tactical decision maker learned from human data. Since many previous works model the intention as a discrete command\cite{codevilla2018end} e.g., turn left and go straight, their motion model has to learn a process actually with larger variance than ours, which is shown in Appendix\ref{variance}.

As one can see, beyond imitating the human driving data, we imitate the design of the conventional hierarchical planning system structure that consists of a local path planner and a trajectory planner, using the driving intention as explicit intermediate representation to decompose the intention and dynamics. From a network training perspective, the potential map is explicitly supervised as a side task \cite{li2018rethinking}, which acts as a semantic regularizer and brings better interpretability.

We validate our method on both datasets and simulator, demonstrating that our method has higher prediction accuracy and generates smoother trajectories.
To summarize, the main contributions of this paper include the following:
\begin{itemize}
	\item We propose a novel representation of trajectory, which brings better driving performance. 
	\item We propose a hierarchical driving model for end-to-end learning from human driving data with explicit intermediate representation, which decompose the intention and dynamics.
	\item Open-loop and closed-loop validation on both datasets and simulator, demonstrating that our method has higher prediction accuracy and generates smoother trajectories. Besides, practical implementation is proposed to tolerate the network latency, which enables generalization testing in real world.
\end{itemize}

\section{Related Work}\label{related-work}
In 2016, benefiting from powerful GPUs and deep neural networks, Bojarski et al. \cite{bojarski2016end} developed \textit{DAVE-2} and tested their method on a real-world platform to prove that deep learning methods can achieve autonomous driving. Although this method can just achieve lane following in simple scenarios, it was a bold start.\par
Recently, a lot of works followed this approach to achieve more intelligent autonomous driving algorithms, including imitation learning methods\cite{codevilla2018end}\cite{codevilla2019exploring} and reinforcement learning methods\cite{liang2018cirl}\cite{wang2020learning}\cite{amini2020learning}.\par

To make the driving models more generic and long-sighted, another paradigm called end-to-end planning is been widely studied. \cite{barnes2017find} \cite{sun2020see} generate paths to guide the vehicle driving, increasing understanding of the environment in driving tasks. \cite{rhinehart2018r2p2} generates diverse paths by learning the multi-mode paths distribution. These methods focus more on the perception of the environment, while the paths generated by these method do not contain dynamic information.\par
\cite{codevilla2018end} \cite{xu2017end} \cite{cai2020vtgnet} generate trajectories with dynamic information, which can be further used for closed-loop tasks. However, most current trajectory generation methods have the following disadvantages: 1) When the controller is actually controlling, it cannot get the trajectory point at any moment, but only take the approximate point, which will reduce the control accuracy. 2) These methods output a fixed number of trajectory points with fixed time intervals. To output more points or other physical quantities, these methods require additional model parameters, which makes these methods data inefficient.\par

Our method corresponds with the end-to-end trajectory planning paradigm, but has some differences from the current methods: 1) Compared with the end-to-end learning methods\cite{yang2018end}\cite{chen2020learning} using RGB images directly, our method using an explicit continuous intention as an intermediate representation\cite{chen2015deepdriving} to decompose the intention (path) and dynamics (trajectory), enhancing model interpretability. 2) Compared with other trajectory generation methods\cite{codevilla2018end} \cite{xu2017end} \cite{cai2020vtgnet} that generate discrete trajectories, we propose a new representation of trajectories, whose contributions are: 1) High-order derivatives can be obtained analytically without adding model parameters, and they can also be used for supervision to achieve better results. 2) A trajectory can be supervised at any sampling interval without adding model parameters, making it sample efficient.

\section{Methodology}

In our previous work\cite{ma2020deepgoal}, the generator network finally obtains the potential map, which is the driving intention that reflects the path information, but without any system dynamic information or temporal information. That means the vehicle only knows where to go, but does not know how to get there. Hence, this approach loses current dynamics information in the training data and does not allow end-to-end learning of human driving trajectories. In this paper, we propose the trajectory generation model, combining with the driving intention module proposed in our previous work\cite{ma2020deepgoal}, so that the whole hierarchical model can be trained end-to-end to learn from human driving trajectories.

\begin{figure*}[t]
\center
\includegraphics[width=0.7\textwidth]{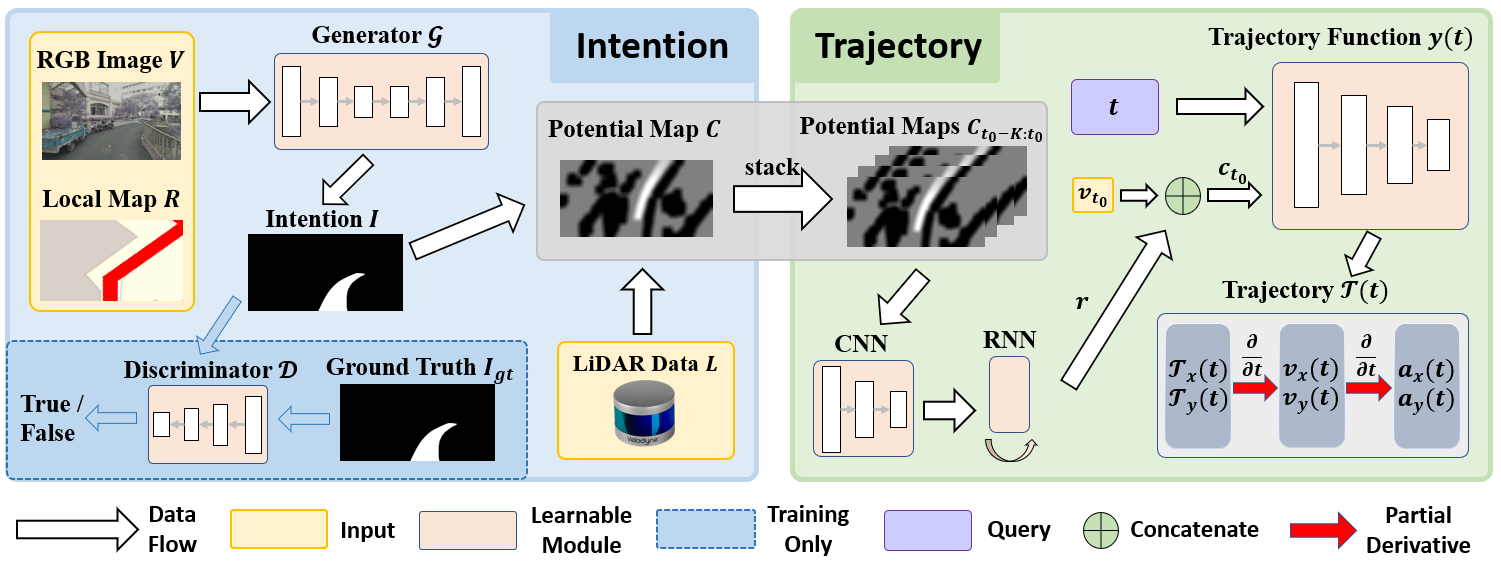}
\caption{
Overview of our hierarchical driving model. The hierarchical driving  model is divided into a \textit{driving intention module} (upper part) and a \textit{trajectory generation module} (lower part). 
The RGB image $V$ is acquired from the vehicle, and local routing planning map $R$ is cropped from an offline map according to the current low cost GPS and IMU data. The intention $I$ is a binary map generated by Generator $\mathcal{G}$ to indicate the local path (white area) to the goal in the current image coordinates, and it is supervised by its ground truth $I_{gt}$ and Discriminator $\mathcal{D}$ which is trained to discriminate whether the input intention is generated by the Generator or is the ground truth. Then the generated intention $I$ is then projected into the top view and fused with LiDAR data $L$ to synthesize the top view grayscale potential map $C$, which is encoded with obstacles and intentions being expressed as grid-based map. The trajectory generation module leverages multi-frame potential maps to extract spatial-temporal feature $r$, and concatenate it with current speed $v_{t_0}$ at time $t_0$ to obtain a condition $c_{t_0}$. Then the trajectory function network $y(t)$ can leverage the condition to generate longitudinal displacement $\mathcal{T}_x$ and lateral displacement $\mathcal{T}_`y$ according to the query time $t$. High-order quantities such as velocity and acceleration can be analytically obtained through partial derivative operation over time $t$, and they form the trajectory $\mathcal{T}$ together with the displacement.
}
\label{fig:structure}
\vspace{-0.5cm}
\end{figure*}

The overall system architecture of our proposed method is shown in Fig.~\ref{fig:structure}, which is divided into a \textit{driving intention module} and a \textit{trajectory generation module}. The driving intention module maps the image $V$ and routing planning $R$ into continuous intention $I$, model it as goal guided potential, and further inject it with obstacle potentials, generating a potential map $C$. Then the trajectory generation module maps a stack of past potential maps $C_{t_0-K:t_0}$, and the current velocity $v_{t_0}$ into a continuous trajectory $\mathcal{T}$ that has high-order smoothness.

\subsection{Driving Intention Module}

The driving intention module is an extension of our previous work \cite{ma2020deepgoal}, which uses an undifferentiable motion planner to generate control, thus blocking the end-to-end learning. We first briefly introduce the generation of continuous intention and the potential map. 

\noindent \textbf{Intention Modeling:} The module takes as input front-view RGB image $V$ combined with routing planning $R$. The routing planning $R$ is generated by usual navigation software, like GoogleMap, and represented as a local crop of the map according to the current position of the vehicle from GPS and IMU. Thus $R$ is a robot-centric representation, which changes its coordinates when the robot moves. For more details please refer to Appendix~\ref{apd:routing-planning}. We model the continuous driving intention $I$ as a mask of $V$ to indicate the local path to the goal in the current image coordinates, which is generated by a U-Net based conditional generator $\mathcal{G}$ with a concatenation of $V$ and $R$ being the input:
\begin{equation}
    I = \mathcal{G}(V, R)
\end{equation}
To build the ground truth $I_{gt}$ of $I$, we refer to the future path executed by the robot in the real world and transform it into the current image coordinates. Therefore, we have a pixel-level supervision for $I$ formulated as a loss term:
\begin{equation}
\mathcal{L}_{int}(\mathcal{G}) = \mathbb{E}_{V, R, I_{gt}}[||I_{gt} - \mathcal{G}(V,R)||_1]
\label{di}
\end{equation}

\noindent \textbf{Adversarial Training:} As the robot only execute in one direction in an intersection in the real world, we only have one-way ground truth for image taken at intersections. For other directions in the intersection, we cannot build (\ref{di}).
To improve the diversity, we use both paired routing planning $R$ and unpaired routing planning $R'$ which is randomly sampled with RGB image $V$, and utilize an adversarial training for supervision.
Specifically, we build a discriminator $\mathcal{D}$ to discriminate whether the driving intention map is generated from the distribution of ground truth data $I_{gt}$, so that exact one-to-one supervision is not required, leading to a conditional adversarial loss term:

\begin{equation}
\begin{aligned}
\mathcal{L}_{intadv}(\mathcal{G},\mathcal{D}) = &\mathbb{E}_{V,R,I_{gt}}[\log \mathcal{D}(V,R,I_{gt})] + \\
&\frac{1}{2}(\mathbb{E}_{R,V}[\log (1-\mathcal{D}(V,R,\mathcal{G}(V,R)))] + \\
&\mathbb{E}_{R',V}[\log (1-\mathcal{D}(V,R',\mathcal{G}(V,R')))])
\end{aligned}
\label{intloss}
\end{equation}

\noindent \textbf{Potential Map Building:} Given the driving intention conditional by the global routing, we actually have goal guidance in robot-centric coordinate. As this mask must lie on the local ground plane, we can transform $I$ into a bird-eye view representation. By smoothing the binary mask, we have a goal guided potential map $C_{goal}$. Furthermore, we represent the perception result as an obstacle potential map $C_{obstacle}$ in the same bird-eye view coordinates. In this paper, we use LiDAR for obstacle detection, but any other obstacle perception is usable. Combining the two potential maps, we have a local potential map $C$ for motion planning, which can be regarded as a local artificial potential field. For more details please refer to Appendix~\ref{apd:potential-map}.

\subsection{Trajectory Generation Module}
The trajectory generation module acts as the motion planner in a conventional pipeline, but it is differentiable, leveraging the end-to-end imitation learning from the human demonstration driving data. The main difficulty for this module is to keep the continuous and smooth nature of the trajectory. Thus the down-stream controller can track the trajectory in a high frequency asynchronously.

\noindent \textbf{Continuous Function Modeling:} For a network with linear decoder, we have a linear mapping from the last hidden layer $f(z)$ to the output $y$ as 
\begin{equation}
y = \sum_i w_i f(z_i) + b
\label{nonact}
\end{equation}
where $w_i$ and $b$ are the network parameters, $f$ is the nonlinear activation function. Furthermore, regarding the input to the network as condition $c$ and a function variable $x$, we have an interpretation of (\ref{nonact}) as
\begin{equation}
y(x) = \sum_i w_i f(z_{i,c}(x)) + b
\label{wlc}
\end{equation}
In this form, the nonlinear activation functions of the last hidden layer form a set of basis functions $f$ parameterized by $z_{i,c}$, deriving a linear combination of basis functions to represent $y(x)$. Compared with the conventional basis functions, the basis functions of (\ref{wlc}) have adaptive forms inferred by the input condition $c$ and $x$. Consequently, this form provides a continuous function approximator, of which the high-order derivatives are naturally derived during the network back-propagation process:
\begin{equation}
\begin{aligned}
\frac{\partial y}{\partial x} &= \sum_i w_i \frac{\partial f}{\partial z_{i,c}}\frac{\partial z_{i,c}}{\partial x}\\
\frac{\partial^2 y}{\partial x^2}& = \sum_i w_i \Bigg ( \frac{\partial f}{\partial z_{i,c}}\frac{\partial^2 z_{i,c}}{\partial x^2} + \frac{\partial^2 f}{\partial z_{i,c} \partial x} \frac{\partial z_{i,c}}{\partial x}^T \Bigg )
\label{dev}
\end{aligned}
\end{equation}
The $2^nd$ order derivative of $y(x)$ is the Hessian matrix. When the dimension of $x$ is high, the computation is intractable. However, if the continuous function is parameterized by low dimensional variable $x$, evaluating the high-order derivatives is feasible.

\noindent \textbf{Trajectory Representation:} At the perspective of continuous function approximator (\ref{wlc}), we therefore use it for trajectory learning. Note that the variable of trajectory $\mathcal{T}$ is the time $t$, we have the high-order derivatives for trajectory i.e., the velocity $v(t)$ and acceleration $a(t)$:
\begin{equation}
\begin{aligned}
&v(t) = \frac{\partial \mathcal{T}(t)}{\partial t} \\
&a(t) = \frac{\partial^2 \mathcal{T}(t)}{\partial t^2}
\label{devobs}
\end{aligned}
\end{equation}
By simply regarding $x$ as $t$, we represent a continuous trajectory $y(t)$ as a neural network with high-order derivatives, and is differentiable for learning. Thanks to the 1-dimensional variable $t$, the evaluation of (\ref{dev}) is very efficient. Moreover, inspired by the conventional Fourier analysis, which is very effective for modeling dynamic system input and output, we also use sinusoidal function\cite{sitzmann2020implicit} as the form of nonlinear activation function $z$, instead of the usual ReLU functions. Finally we have the neural trajectory as
\begin{equation}
y(t) = \sum_i w_i \cos(z_{i,c}(t)) + b
\label{sinwlc}
\end{equation}
This function can be supervised by the any order of derivatives to learn the parameters\cite{sitzmann2020implicit}. In this paper, at time $t_0$, we build loss terms based on the observed position $\hat{\mathcal{T}}(k)$, velocity $\hat{v}(k)$ and the acceleration $\hat{a}(k)$ at several sampling time $k$:
\begin{equation}
\begin{aligned}
\mathcal{L}_{\mathcal{T}}(y) =& \mathbb{E}_{t_0}[ \sum_{k\in\{t_0,t_0+T\}} ||\hat{\mathcal{T}}(k) - y(k)||_2^2\\&+ 
\lambda_1 ||\hat{v}(k) - \frac{\partial y(t)}{\partial t}|_{t=k}||_2^2 \\ &+ \lambda_2 ||\hat{a}(k) - \frac{\partial^2 y(t)}{\partial t^2}|_{t=k}||_2^2]
\end{aligned}
\label{tcost}
\end{equation}
where $T$ is the time horizon for the trajectory. 
 
We briefly summarize the advantage of modeling continuous trajectory as the \textit{smoothness} for execution, which is guaranteed by the infinitely order differentiable sinusoidal basis functions, and the \textit{minimal parameterization} for compatible position, velocity, and acceleration generation, which is guaranteed by the continuous function approximator when regarding the last layer of the network as a linear combination to the output.

\noindent \textbf{Conditional Generation:} To embed the proposed differentiable representation into the trajectory generation module, we need to specify the condition $c$. In driving scenarios, the factors we consider for trajectory planning include the ego-vehicle dynamics, the surrounding obstacles dynamics, as well as the stationery environmental obstacles. For the first factor, we include the current velocity $v_{t_0}$, as one of the condition. For the second and third factors, since the potential map $C$ only reflects a cut of the surrounding obstacles, we encode a window of multiple previous potential maps $C_{t_0-K:t_0}$ via a CNN and a RNN, denoted as $r$. Resultantly, we have a condition at $t_0$ as
\begin{equation}
c_{t_0} = \left[\begin{array}{cc}
v_{t_0} & r(C_{t_0-K:t_0})
\end{array}  \right]^T
\label{cond}
\end{equation}
which becomes a mapping from $t$ to the frequency and phase of the sinusoidal basis function. Then we can derive the trajectory from $t_0$ to $t_0+T$ by sweeping $t$ from $t_0$ to $t_0+T$.

\subsection{Loss Function Design and Training}
Based on the loss terms design introduced above, we can simply set the loss function for training the network as a sum of (\ref{di}), (\ref{intloss}) and (\ref{tcost}):
\begin{equation}
\mathcal{L} = \mathcal{L}_{intadv}(\mathcal{G},\mathcal{D})+\mathcal{L}_{int}(\mathcal{G})+\mathcal{L}_{\mathcal{T}}(y)
\label{loss}
\end{equation}

\noindent \textbf{Data Relabeling:} When predicting for the future trajectory, there can be non-causal labeling for the current data. For example, a vehicle stops in a queue, waits for the traffic light turning green. Say, there are $3s$ left for the light turning. Then the retrospective trajectory label with a time horizon $5s$ for the current time is: stop for $0-3s$ and accelerate for $3s-5s$. Therefore, the supervision becomes a trajectory crossing the preceding vehicle. However, the driver at the current time actually cannot know that the light turns in $3s$, hence such supervision is reasonable unless the driver was informed with an external aid. To relieve this labeling problem, we re-label the supervision for such crossing cases with causality by modifying the trajectory with deceleration at a constant acceleration before the collision happens. We consider it is reasonable behavior for human driving without referring to the future.

\noindent \textbf{Training Details:} 
We use a batch size of $32$ and Adam optimizer with an initial learning rate of $0.0003$. The networks are implemented in PyTorch and trained on AMD 3900X CPU and Nvidia RTX 2060 Super GPU until the model converges. For more details about the network structure, please refer to Appendix~\ref{apd:network}.

\subsection{Closed-loop System Implementation}

\begin{figure}[t]
\center
\includegraphics[width=0.4\textwidth]{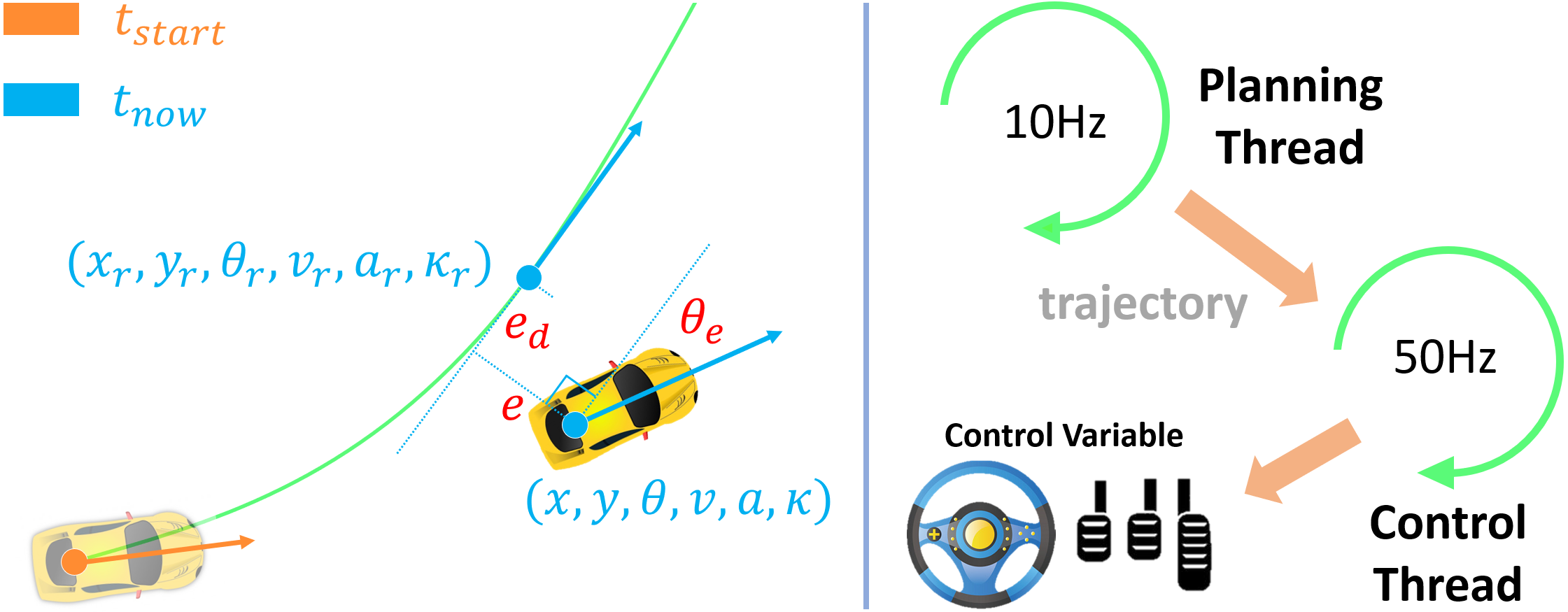}
\caption{Asynchronous planning and control. The planning thread generates a trajectory at $t_{start}$, and the control thread tracks the whole trajectory w.r.t. time. At time $t_{now}$, the lateral error $e$ and heading error $\theta_e$, together with curvature $\kappa_r$ are fed into a steering controller. The throttle and brake command are computed by PID controller according to the difference between $v_r$ and current $v$.}
\label{ctrl}
\vspace{-0.3cm}
\end{figure}

Different from the works using only a point position for tracking, which may not perform well due to the limited onboard computing frequency, we implement an asynchronous driving architecture by separating the long-horizon trajectory planning and trajectory tracking in two threads, achieving a receding horizon controller\cite{watterson2015safe}, to overcome the time-delay issue, which is shown in Fig.~\ref{ctrl}.

\noindent \textbf{Controller design:} Specifically, the planning thread generates continuous trajectories with time horizon $T=3s$ in a fixed frequency or triggered, and annotates the trajectory with vehicle's pose and the starting time $t_{start}$. The control thread is running in another fixed higher frequency. It gets vehicle's current pose at time $t_{now}$ and queries the trajectory with $t = t_{now}-t_{start}$ as the reference point. Then we use a feedback controller for throttle and brake control, and a nonlinear rear wheel feedback controller \cite{paden2016survey} for steering control. For more details about the controller, please refer to Appendix~\ref{apd:ctrl}.

\section{Experiments}
\subsection{Open-loop Experiments on Datasets}
We first validate the effectiveness of our proposed method on two publicly available and widely recognized datasets: KITTI Raw Data \cite{geiger2013vision} and Oxford Radar RobotCar \cite{barnes2020oxford}.

\noindent \textbf{Datasets:} \textit{KITTI Raw Data (KITTI)} contains $6$ hours of traffic scenarios at $10$ Hz using a variety of sensor modalities including high-resolution stereo cameras, a Velodyne 3D laser scanner, and a high-precision GPS/IMU inertial navigation system (INS).
\textit{Oxford Radar RobotCar (RobotCar)} is a radar extension to The Oxford RobotCar Dataset, providing data from Dual Velodyne HDL-32E LIDARs and one Bumblebee XB3 trinocular stereo camera with optimized ground truth visual odometry for $280~km$ of driving around Oxford University.

\noindent \textbf{Implementation Details:} Since there is no routing planning $R$, intention image $I$ and potential map $C$ in these datasets directly, we first build them up before training. For KITTI, we extract a time-independent $30m$ path from INS which is then projected into image coordinate by perspective mapping to generate intention map $I$. Potential map $C$ is acquired by fusing point cloud and reprojected generated intention image $I$ by driving intention module using inverse perspective mapping. Note that the point cloud data is first transformed into INS coordinate, ahead of fusion, with calibration information between Velodyne and INS. As for Oxford, the data preparation process is similar to KITTI, with the exception that we use visual odometry (VO) data as ground truth path due to poor GPS signals Oxford provides. Additionally, we undistort image data collected by the stereo camera according to \cite{barnes2020oxford}. For both datasets, we split them as $70\%$ are used for training with $10\%$ and $20\%$ for evaluation and testing.

\noindent \textbf{Evaluation Metrics:} We use six evaluation metrics to assess the effectiveness of various models, all of which are computed over the testing part of datasets.
\begin{itemize}
	\item Average Displacement Error (ADE, $\mathcal{E}_{ad}$) \cite{cai2020vtgnet}: Average L2 distance between the ground truth and generated paths over all positions. Note that the corresponding positions are found w.r.t time rather than the nearest neighbor.
	\item Final Displacement Error (FDE, $\mathcal{E}_{fd}$) \cite{cai2020vtgnet}: L2 distance between the last ground truth and generated positions.
	\item Average Longitudinal Error $\mathcal{E}_{x}$, Average Lateral Error $\mathcal{E}_{y}$, and Average Velocity Error $\mathcal{E}_{v}$ \cite{cai2020vtgnet} represent mean longitudinal, lateral displacement and velocity scalar error between the ground truth and the corresponding quantities estimated from the generated trajectory respectively. Among them, $\mathcal{E}_{v}$ can measure the model's ability to model system dynamics.
	\item Smoothness of the trajectory, which is average jerk\cite{fan2018baidu} of the trajectory $\mathcal{S}$.
\end{itemize}

\noindent \textbf{Ablation Study:} We exhibit the principle of utilizing our method by defining five ablated models from the original model as below:
\begin{itemize}
	\item w/o intention: we skip the intention module and use raw RGB images as input instead of potential maps;
	\item w/o v0: we remove the initial velocity input to the model;
	\item w/o cos: replace sinusoidal function to ReLU;
	\item w/o HOS: the model has no high-order supervision;
	\item w big HOS: big weights ($\lambda_1=1.0,\lambda_2=0.5$) of high-order supervision in the loss function;
	\item Ours: normal weights ($\lambda_1=0.2,\lambda_2=0.05$) of high-order supervision in the loss function.
\end{itemize}

We refer to Tab.~\ref{ablation-studies} for checking the influence of different models on error metrics. When some stimuli are missing, performance drops significantly, which validates the importance of intention. The performance degenerates most is the model w/o v0, because the initial velocity $v_0$ reflects the current dynamics, without which the trajectory does not have boundary value condition. Results on w/o HOD and big HOD indicate that the trade-off between different order supervisions is indispensable. On the one hand, high-order supervision rectifies dynamic outputs of Neural Trajectory. w/o cos verifies a better approximation accuracy when basis function is selected more appropriate for the dynamic system.

\begin{table}[t]
\center
\caption{Ablation Study on KITTI Dataset}
\label{ablation-studies}
\resizebox{200pt}{!}{\begin{tabular}{c@{\ }c@{\ }c@{\ }c@{\ }c@{\ }c@{\ }c}
\toprule
Algorithm & $\mathcal{E}_{ad}$ & $\mathcal{E}_{x}$ & $\mathcal{E}_{y}$ & $\mathcal{E}_{fd}$& $\mathcal{E}_{v}$& $\mathcal{S}$\\
& $(m)$ & $(m)$ & $(m)$ & $(m)$& $(m/s)$& $(m/s^3)$\\
\midrule
w/o intention & $2.65$ & $2.15$ & $0.85$ & $3.35$ & $2.24$ & $0.36$\\
w/o v0 & $1.75$ & $1.36$ & $0.57$ & $2.65$ & $2.18$ & $0.30$\\
w/o cos & $1.04$ & $0.89$ & $0.43$ & $1.80$ & $0.96$ & $0.50$\\
w/o HOS & $2.02$ & $1.62$ & $0.79$ & $2.80$ & $1.78$ & $0.64$\\
w big HOS & $1.88$ & $1.47$ & $0.56$ & $2.25$ & $0.89$ & $\mathbf{0.26}$\\
$\mathbf{Ours}$ & $\mathbf{0.99}$ & $\mathbf{0.80}$ & $\mathbf{0.40}$ & $\mathbf{1.64}$ & $\mathbf{0.88}$& $0.28$\\
\bottomrule
\end{tabular}}
\vspace{-0.2cm}
\end{table}

\noindent \textbf{Comparative Study:} For comparison, we investigate three recent baseline methods, and these models are introduced as follows:
\begin{itemize}
	\item NI+DT and NI+DT+RNN: No intention (NI) input and discrete trajectory (DT) output. We follow the method in \cite{xu2017end}. In NI+DT, the CNN module intends to extract features from sequential RGB images inputs, and the concatenated features are fed into fully connected layers to generate a discrete trajectory. In NI+DT+RNN, the network runs analogical to NI+DT except that the extracted features are fused by \textit{Long Short-Term Memory (LSTM)} instead of simply concatenating them.
	\item DI+DT: Discrete intention (DI) input and discrete trajectory (DT) output. We follow the idea of \cite{codevilla2018end} to give three discrete commands as discrete intentions to switch three different networks. Instead of generating control commands, we change it to output a series of discrete trajectory points and velocities.
	\item CI+DT and CI+DT+RNN: Continuous intention (CI) input and discrete trajectory (DT) output. The same structure to NI+DT and NI+DT+RNN, but we replace sequential RGB images inputs to sequential potential maps as continuous intention.
	\item VTGNet\cite{cai2020vtgnet}: SOTA method we compared, which uses discrete intents and generates discrete trajectories
	\item CI+DWA: The trajectory generation method used in our previous work\cite{ma2020deepgoal}. We keep the pre-trained driving intention module and use the traditional dynamic window approach (DWA) to generate simple trajectories with fixed velocity and angular velocity to avoid obstacles and navigate to the destination.
	\item CI+Ploy: A learning-based parametric model that models trajectories as polynomials. For fairly comparation, we keep the driving intention module and only modify the trajectory generation module propose in our method by removing time $t$ from the input and changing the output to an $K^{th}$-order polynomial coefficients. We take $K=5$\cite{fan2018baidu} for the experiment.
	\item CI+CT (Ours): Continuous intention (CI) input and continuous trajectory (CT) output method that we proposed.
\end{itemize}

From the results shown in Tab.~\ref{dataset-result}, we can clearly see that our network performs the best on all predictive metrics, which means the trajectories our method generates are smooth and more accurate in accordance with displacement-level metrics such as $\mathcal{E}_{ad}$ and high-order metrics like $\mathcal{E}_{v}$. Tab.~\ref{dataset-result} shows that DI+DT outperforms NI+DT and NI+DT+RNN, which indicates that the superiority of introducing intention. The intention helps to guide the direction of trajectory generation, resulting in the improvement of zero-order metrics. Additionally, model performance is enhanced by introducing continuous intention rather than discrete commands. Most importantly, our model significantly decreases the high-order metrics on both KITTI and RobotCar. Thanks to the continuous representation, we manage to precisely model the smoothness and the intrinsic constraints between derivatives. For more details about the controller, please refer to Appendix~\ref{apd:ctrl}.
We also prove that our trajectory generation method has better prediction results than the polynomial method shown in Tab.~\ref{dataset-result}. We believe that the polynomial trajectory generation method introduces two types of errors: 1) Fitting error from approximating trajectories with polynomials. 2) Prediction error due to prediction polynomial coefficients. While our method only has prediction error, which can reduce uncertainty of our model. This proves the superiority of our method over parametric polynomial models for trajectory prediction. And it also proves the superiority of continuous trajectory generation methods, comparing with discrete trajectory generation methods.

\begin{table}[t]
\center
\caption{Comparative Study Results on Two Benchmark Datasets}
\label{dataset-result}
\resizebox{240pt}{!}{\begin{tabular}{l@{\ \ }c@{\ \ }c@{\ \ }c@{\ \ }c@{\ \ }c@{\ \ }c@{\ \ }c}
\toprule
Dataset & Algorithm & $\mathcal{E}_{ad}$ & $\mathcal{E}_{x}$ & $\mathcal{E}_{y}$ & $\mathcal{E}_{fd}$& $\mathcal{E}_{v}$ & $\mathcal{S}$\\
  &   & $(m)$ & $(m)$ & $(m)$ & $(m)$& $(m/s)$ & $(m/s^3)$\\
\midrule
\multirow{11}{*}{KITTI} & NI+DT~\cite{xu2017end} & $4.91$ & $4.53$ & $1.11$ & $8.54$ & $3.88$& $0.96$\\
& NI+DT+RNN~\cite{xu2017end} & $4.85$ & $4.46$ & $0.90$ & $8.44$ & $3.79$& $0.87$\\
& DI+DT~\cite{codevilla2018end} & $4.43$ & $3.76$ & $1.30$ & $6.72$ & $3.21$& $0.70$\\
& DI+CT & $2.60$ & $2.00$ & $0.51$ & $2.55$ & $1.68$& $0.39$\\
& CI+DT & $3.96$ & $3.76$ & $0.56$ & $7.27$ & $3.96$& $0.87$\\
& CI+DT+RNN & $3.36$ & $3.20$ & $0.42$ & $6.00$ & $2.31$& $0.79$\\
& VTGNet\cite{cai2020vtgnet} & $2.98$ & $2.51$ & $0.51$ & $4.89$ & $1.48$& $0.58$\\
& CI+DWA\cite{ma2020deepgoal} & $2.27$ & $1.55$ & $1.29$ & $3.80$ & $2.17$& $5.13$\\
& CI+Poly & $1.22$ & $1.03$ & $0.44$ & $2.46$ & $1.27$& $0.33$\\
& $\mathbf{CI+CT (Ours)}$ & $\mathbf{0.99}$ & $\mathbf{0.80}$ & $\mathbf{0.40}$ & $\mathbf{1.64}$ & $\mathbf{0.88}$& $\mathbf{0.28}$\\
\midrule
\multirow{11}{*}{RobotCar} & NI+DT~\cite{xu2017end} & $2.83$ & $2.81$ & $0.33$ & $5.12$ & $2.02$& $0.23$\\
& NI+DT+RNN~\cite{xu2017end} & $2.58$ & $2.46$ & $0.25$ & $4.89$ & $1.77$& $0.83$\\
& DI+DT~\cite{codevilla2018end} & $2.31$ & $2.08$ & $0.56$ & $4.06$ & $1.66$& $0.78$\\
& DI+CT & $1.51$ & $1.00$ & $0.46$ & $1.58$ & $0.84$& $0.23$\\
& CI+DT & $1.76$ & $1.66$ & $0.35$ & $3.33$ & $1.32$& $0.41$\\
& CI+DT+RNN & $1.75$ & $1.56$ & $0.37$ & $3.15$ & $1.41$& $0.60$\\
& VTGNet\cite{cai2020vtgnet} & $0.88$ & $0.78$ & $0.28$ & $1.77$ & $0.56$& $0.79$\\
& CI+DWA\cite{ma2020deepgoal}\textbf{} & $0.83$ & $0.71$ & $0.36$ & $1.78$ & $0.73$& $4.70$\\
& CI+Poly & $0.82$ & $0.69$ & $0.30$ & $1.87$ & $0.72$& $0.24$\\
& $\mathbf{CI+CT (Ours)}$ & $\mathbf{0.76}$ & $\mathbf{0.55}$ & $\mathbf{0.26}$ & $\mathbf{1.43}$ & $\mathbf{0.42}$& $\mathbf{0.16}$\\
\bottomrule
\end{tabular}}
\vspace{-0.3cm}
\end{table}

\subsection{Closed-loop Experiments in Simulation}
\begin{table*}[htbp]
\vspace{-0.3cm}
\caption{CARLA Benchmark Evaluation Results on Success Rate ($\%$)}
\label{carla-result}
\begin{threeparttable} 
\begin{tabular}{l@{\ }c@{\ \ \ \ }c@{\ \ \ \ }c@{\ \ \ \ }c@{\ \ \ \ }c@{\ \ \ \ }c@{\ \ \ \ }c@{\ \ \ \ }c@{\ \ \ \ }c}
\toprule
    Task & Condition              & MT~\cite{li2018rethinking} & CIL~\cite{codevilla2018end} & CIRL~\cite{liang2018cirl} & CAL~\cite{sauer2018conditional} & CILRS~\cite{codevilla2019exploring} & LSD\cite{ohn2020learning} & LaTeS\cite{zhao2019lates} & $\mathbf{Ours^\dagger}$\\
    \midrule
Straight & \multirow{4}{*}{Train} & $96$ & $98$ & $98$ & $\mathbf{100}$& $96$ & -     & -     & $\mathbf{100}$\\
One turn &                        & $87$ & $89$ & $97$ & $97$ & $92$ & -     & -     & $\mathbf{100}$\\
Navigation &                      & $81$ & $86$ & $93$ & $92$ & $95$ & -     & $\mathbf{100}$ & $\mathbf{100}$\\
Nav. dynamic &                    & $81$ & $83$ & $82$ & $83$ & $92$ & -     & $\mathbf{100}$ & $\mathbf{100}$\\
\midrule
Straight & \multirow{4}{*}{New Town \& New Weather}  & $96$ & $80$ & $98$ & $94$ & $96$ & $\mathbf{100}$ & -     & $\mathbf{100}$\\
One turn &                        & $82$ & $48$ & $80$ & $72$ & $92$ & $\mathbf{100}$ & -     & $\mathbf{100}$\\
Navigation &                      & $78$ & $44$ & $68$ & $68$ & $92$ & $98$  & $98$  & $\mathbf{100}$\\
Nav. dynamic &                    & $62$ & $42$ & $62$ & $64$ & $90$ & $92$  & $\mathbf{98}$  & $96$\\
\bottomrule
\end{tabular}
\begin{tablenotes}
    \footnotesize
    \item Adding $\dagger$ denotes the model trained and evaluated on CARLA 0.9.9.4, while not adding it denotes that the model is trained and evaluated on CARLA $\le$ 0.9.5.
  \end{tablenotes}
\end{threeparttable}
\vspace{-0.4cm}
\end{table*}

As a driving task, only open-loop validation is not sufficient, since the error is accumulated. To further validate the efficacy of our proposed method, we evaluate our method with closed-loop experiments both in the CARLA simulation and a real-world vehicle platform.

\noindent \textbf{Experiments Setup:} Since we need to obtain highly interpretable and smooth trajectories as expert-provided data to imitate, we use human driving data as our training data rather than built-in AI driving data. We use Logitech G29 driving force-racing wheel as our data acquisition equipment, and collect about 2 hours of human driving data in CARLA simulation with a speed limit of $30~km/h$. We collect data in Town01 with 4 different weather as \textit{training condition} and test our method in Town02 with other 2 different weather as \textit{testing condition}.

We compare our method performance with several previously proposed approaches \cite{li2018rethinking, codevilla2018end, liang2018cirl, sauer2018conditional, codevilla2019exploring, ohn2020learning, zhao2019lates, prakash2020exploring} on the original CARLA benchmark \cite{dosovitskiy2017carla} and the NoCrash benchmark \cite{codevilla2019exploring}. The original CARLA benchmark allows us to compare algorithms on sets of strictly defined goal-directed navigation tasks. It provides 4 simple navigation tasks, the last of which has a few dynamic obstacles. The NoCrash benchmark is much more challenging. The vehicle drives in 3 different traffic conditions: \textit{empty town} where no dynamic objects exist, \textit{regular traffic} which has moderate number of cars and pedestrians, and \textit{dense traffic} with large number of pedestrians and heavy traffic condition. Almost all current methods have a low success rate under the most difficult dense condition under testing condition.

\noindent \textbf{Comparative Study:} The comparison results on CARLA benchmark are shown in Tab.~\ref{carla-result}. We achieve a $100\%$ success rate on all tasks under training conditions and 3 tasks under testing conditions, outperforming all other methods. On the last task under testing condition, our success rate is slightly lower than that of LaTeS. Overall, our method achieves competitive performance with state-of-the-art methods.

\begin{figure}[tb]
\center
\includegraphics[width=0.45\textwidth]{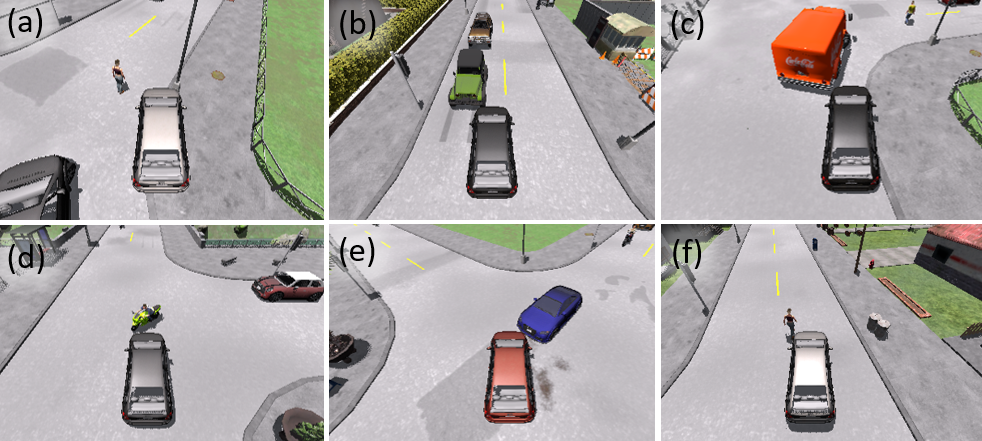}
\vspace{-0.2cm}
\caption{Representative failure cases among all methods. (a) Hit a static obstacle; (b) Crash into wrong side of road; (c) Hit a car from rear side; (d-f) Hit moving agents at side. Most of our failure cases are (d-f).}
\label{fail}
\vspace{-0.2cm}
\end{figure}

We test our method 3 times on NoCrash benchmark and the results are shown in Tab.~\ref{carla-result2}. In empty condition task, our method achieves a $100\%$ success rate both in training and testing conditions. In regular traffic task, our method is relatively $35\%$ higher than the second place in testing condition and equals to LaTeS in training condition. In dense traffic task, our method is relatively $35\%$ and $91\%$ higher than the second place in training and testing. Several common and representative failure cases in all methods are shown in Fig.~\ref{fail}. In our test, there are very few cases of (a), and (b), which indicates the superiority of introducing continuous intention and long-term planning, and there are also few cases of (c), which indicates our method learns the system dynamics better. Almost all failure cases are collisions with side-on vehicles and pedestrians at intersections in our test, as shown in Fig.~\ref{fail} (d-f). All methods encounter this problem. One reason is that agents in CARLA simulation are not strong enough to avoid obstacles at intersections. And we suspect that there is less data for this scenario in our training data and thus our model is less concerned with dynamic obstacles that are not in the driving intention area.

\makeatletter\def\captype{table}\makeatother
\begin{table}[tb]
    \setlength{\abovecaptionskip}{-0.2cm}
    \setlength{\belowcaptionskip}{0.cm}
	\caption{NoCrash Benchmark Evaluation Results on Success Rate ($\%$)}
	\label{carla-result2}
	\begin{threeparttable}
	\begin{center}
	    \resizebox{200pt}{!}{
		\begin{tabular}{lcccccc}
			\toprule
			& \multicolumn{3}{c}{Training Condition}  & \multicolumn{3}{c}{New Town \& New Weather} \\
			Method             & Empty            & Regular           & Dense            & Empty            & Regular           & Dense \\ 
			\midrule
			CIL~\cite{codevilla2018end}     & \ \ $79\pm1$     & $60\pm1$          & $21\pm2$         & $24\pm1$         & $13\pm2$          & \ $2\pm0$ \\
			CAL~\cite{sauer2018conditional}     & \ \ $81\pm1$     & $73\pm2$          & $42\pm3$         & $25\pm3$         & $14\pm2$          & $10\pm0$ \\
			MT~\cite{li2018rethinking}       & \ \ $84\pm1$     & $54\pm2$          & $13\pm4$         & $57\pm0$         & $32\pm2$          & $14\pm2$\\
			CILRS~\cite{codevilla2019exploring} & \ \ $97\pm2$     & $83\pm0$          & $42\pm2$         & $90\pm2$         & $56\pm2$          & $24\pm8$\\
			LSD+~\cite{ohn2020learning}    & \ \ -            & -                 & -                & $95\pm1$         & $65\pm4$          & $32\pm3$\\
			LaTeS~\cite{zhao2019lates} & \ \ $\mathbf{100}\pm\mathbf{0}$    &$\mathbf{94}\pm\mathbf{2}$           & $54\pm3$         & $83\pm1$         & $68\pm7$          & $29\pm2$\\
			DA-RB~\cite{prakash2020exploring} & \ -        & -                & $66\pm5$          & -                & -                 & $35\pm2$\\
			DI+CT$^\dagger$ & \ \ $97\pm3$     & $92\pm2$          & $73\pm2$         & $91\pm2$         & $70\pm2$          & $56\pm3$\\
			CI+DT$^\dagger$ & \ \ $97\pm2$     & $83\pm3$          & $60\pm3$         & $93\pm1$         & $71\pm2$          & $50\pm3$\\
			$\mathbf{Ours^\dagger}$        & \ \ $\mathbf{100}\pm\mathbf{0}$    & $\mathbf{94}\pm\mathbf{2}$          & $\mathbf{89}\pm\mathbf{3}$         & $\mathbf{100}\pm\mathbf{0}$        & $\mathbf{92}\pm\mathbf{1}$         & $\mathbf{67}\pm\mathbf{2}$\\
			\bottomrule
		\end{tabular}}
	\end{center}
\begin{tablenotes}
    \footnotesize
    \item Adding $\dagger$ denotes the model trained and evaluated on CARLA 0.9.9.4, \\
    while not adding it denotes that the model is trained and evaluated \\
    on CARLA $\le$ 0.9.5.\\
  \end{tablenotes}
\end{threeparttable}
\vspace{-0.6cm}
\end{table}

\noindent \textbf{Generalization to other vehicle:} Similar to \cite{cai2020vtgnet}, we also transfer our method from cars to motorcycles to validate the generalization ability for different vehicles of our method, because motorcycles have smaller turning radius, greater acceleration, and its camera view tilts and shifts when turning and braking. We test in the regular task and training condition on NoCrash benchmark, and we even do not change anything including controller parameters. The success rate is $91\%$, which is only $3\%$ below the original $94\%$ and validates the strong generalization ability of our method.

\begin{figure}[t]
\center
\includegraphics[width=0.3\textwidth]{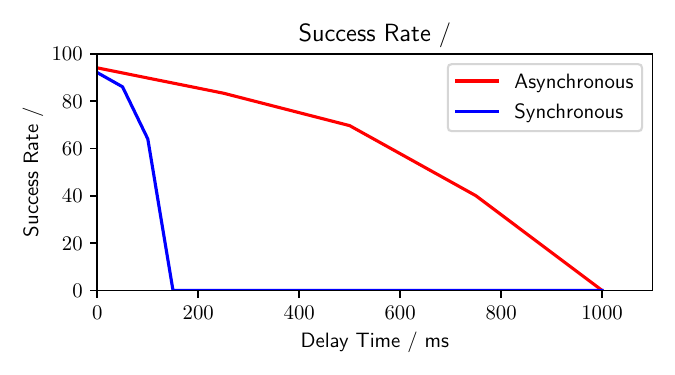}
\vspace{-0.45cm}
\caption{Success rate under different time delays.}
\label{delay}
\end{figure}

\noindent \textbf{Robustness against computation latency:} We also evaluate our method in different time delays to test the ability to eliminate the impact due to latency, comparing with synchronous driving architecture which is the built-in autonomous driving AI in CARLA simulation. To achieve this, we artificially add different time delays to the planning process in the regular task and training condition on NoCrash benchmark. The result is shown in Fig.~\ref{delay}. Our method still has a comparable success rate when the latency is up to $500~ms$, which validates the robustness against the computation latency of our method.

\begin{figure}[tb]
\vspace{-0.05cm}
\center
\includegraphics[width=0.4\textwidth]{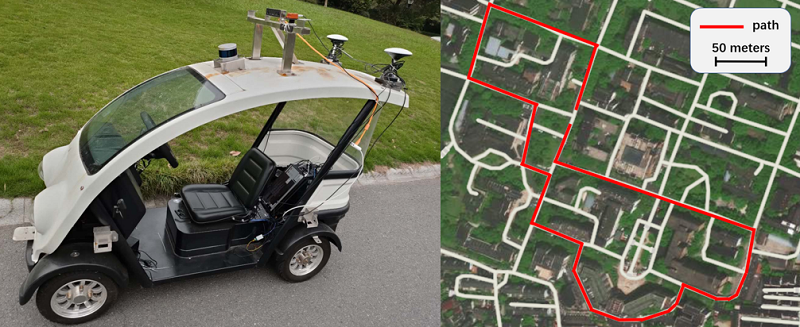}
\caption{Real-world vehicle platform and the actual travel route in the map.}
\label{vehicle-map}
\vspace{-0.3cm}
\end{figure}

\subsection{Closed-loop Experiments in Real World}
We use a real-world vehicle with Ackermann steering for experiments in a campus environment, shown in Fig.~\ref{vehicle-map}. The vehicle is equipped with a Velodyne VLP-16 LiDAR, a MYNT EYE D-1000-120 camera, an Xsens MTi-300 IMU, a Qianxun D300-GNSS GPS module, and an industrial PC with Intel i7-7700 CPU which collects all sensors' data, runs our method, and sends control signals to the vehicle by CAN bus. We build a route map of campus and plan a global route to guide the vehicle. We train our model and VTGNet\cite{cai2020vtgnet} with real-world data, and test the two models in real-world closed-loop experiment, counting the number of human interruptions. The result is that VTGNet has 8 human interruptions occur totally, while our method has 5 human interruptions occur totally, which proves the superiority of our method.\\
We also use another dataset YQ21\footnote{\url{https://tangli.site/projects/academic/yq21/}} for testing generalization ability of our method and VTGNet\cite{cai2020vtgnet}. We directly test the models on YQ21, without re-training the models. The experimental result also shows our method has stronger generalization ability. For more experimental results, please refer to Appendix~\ref{apd:generalization}.



\section{Conclusion}\label{conclusion}
In this paper, we propose a hierarchical driving model with explicit intermediate representation to decomposes the intention (path) and dynamics (trajectory), bringing better interpretability and without losing advantages of end-to-end training. IIn addition, we propose a new representation of trajectory which is continuous and differentiable. However, our method has still some limitations: 1) Since our method is still imitation learning or called behavior cloning, it is difficult for our method to learn good causality from training data. 2)Similar to many current trajectory generation methods, our approach does not consider lanes, traffic lights and traffic rules. We hope to address these issues in our future work.






\bibliographystyle{ieeetr}
\bibliography{reference}

\clearpage
\begin{appendices}
\renewcommand{\thetable}{A\arabic{table}}
\renewcommand{\thefigure}{A\arabic{figure}}
\renewcommand{\theequation}{\thesection.\arabic{equation}}
\setcounter{table}{0}
\setcounter{figure}{0}

\section{Data Variance}\label{variance}
\begin{figure*}[htbp]
\centering
\subfigure[Trajectories in Dataset]{
\begin{minipage}[htbp]{0.32\linewidth}
\centering
\includegraphics[width=\textwidth]{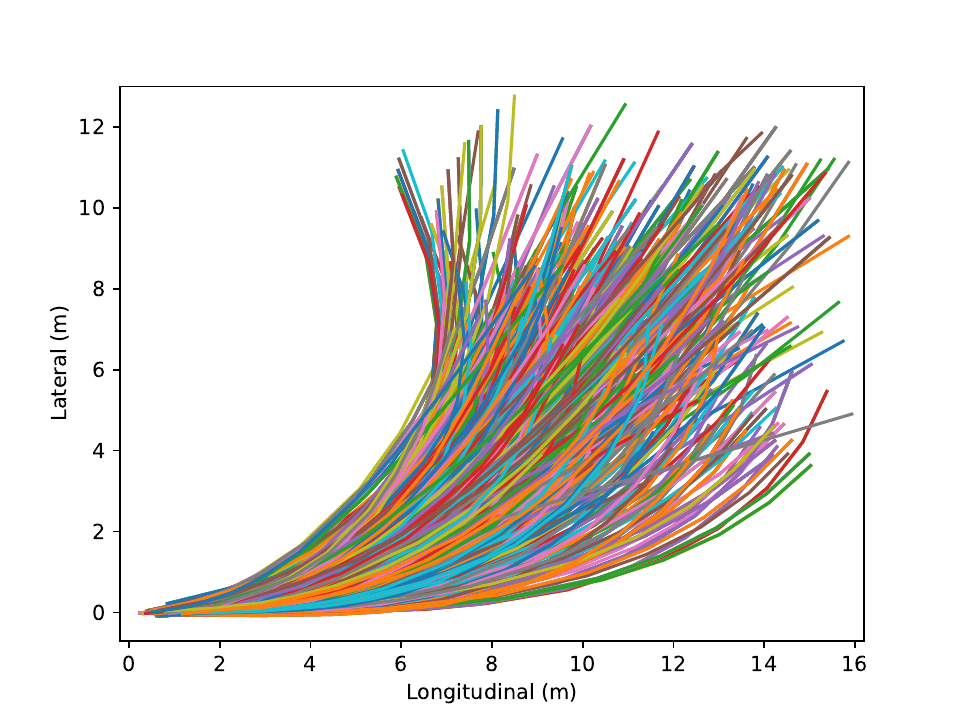}
\label{org-left}
\end{minipage}%
}%
\subfigure[Trajectories Generated by Discrete Intention Method]{
\begin{minipage}[htbp]{0.32\linewidth}
\centering
\includegraphics[width=\textwidth]{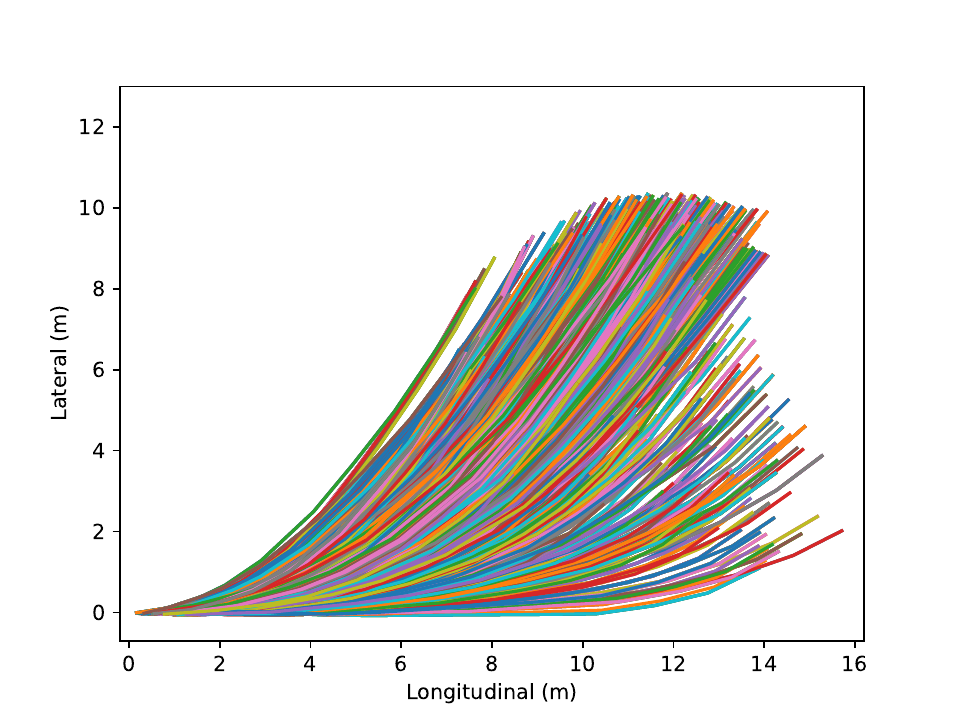}
\label{di-left}
\end{minipage}%
}%
\subfigure[Trajectories Generated by Continuous Intention Method]{
\begin{minipage}[htbp]{0.32\linewidth}
\centering
\includegraphics[width=\textwidth]{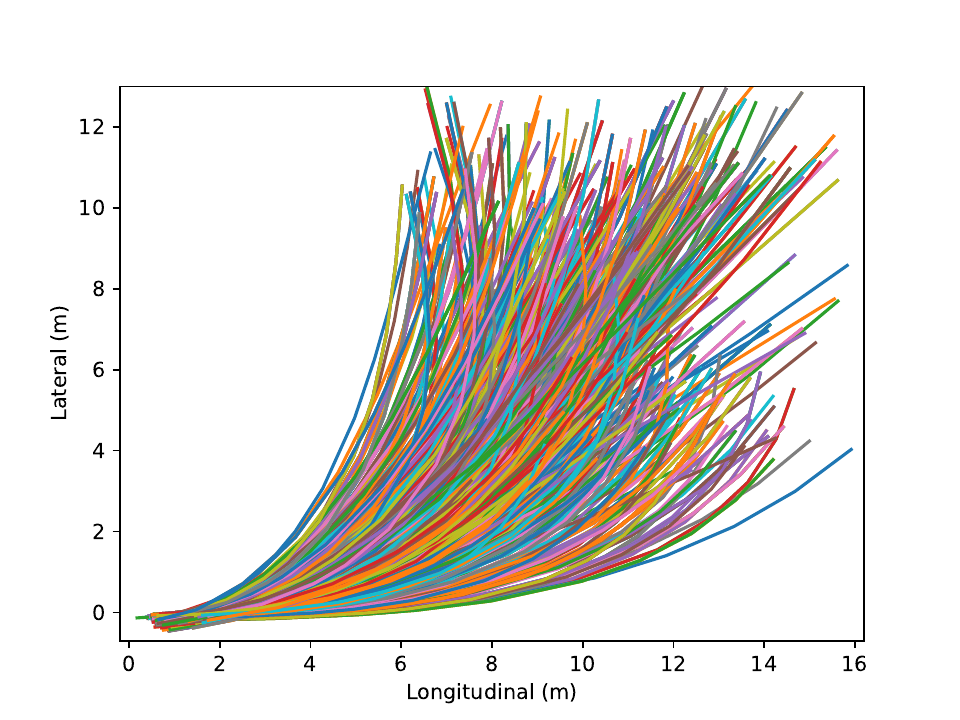}
\label{ci-left}
\end{minipage}
}%
\centering
\caption{Trajectories Distribution with Left Turn Command.}
\end{figure*}

In the approach using discrete intention, one intention corresponds to multiple different trajectories, and the variability between the different trajectories may be large, which is shown in Fig. A\ref{org-left}, where multiple different trajectories from KITTI dataset corresponding to the left turn command. Thus, for a specific intention, the range of trajectories that need to be learned is large, and the data itself has a large variance. By introducing continuous driving intention, our method learns data with less variance, which is benefit to model learning.

In Fig. A\ref{di-left}, it shows trajectories generated by the discrete intention method with left turn command, and Fig. A\ref{ci-left} shows those generated by our continuous intention method.
It seems that trajectories generated by the discrete intention method have a high degree of similarity, and some trajectories with large curvature are not generated. That means the discrete intention method does not learn a good distribution of trajectories with high variance. And it also shows that our continuous intention method can generate diverse trajectories, and the distribution of trajectories is more in line with the distribution of trajectories in the dataset which is shown in Fig. A\ref{org-left}.

\section{Definition of Routing Planning Maps}\label{apd:routing-planning}
\begin{figure}[htbp]
\centering
\subfigure[Full Offline Routing Planning. The blue box represents the area cropped according to GPS and yaw angle.]{
\begin{minipage}[htbp]{0.5\linewidth}
\centering
\includegraphics[width=0.6\textwidth]{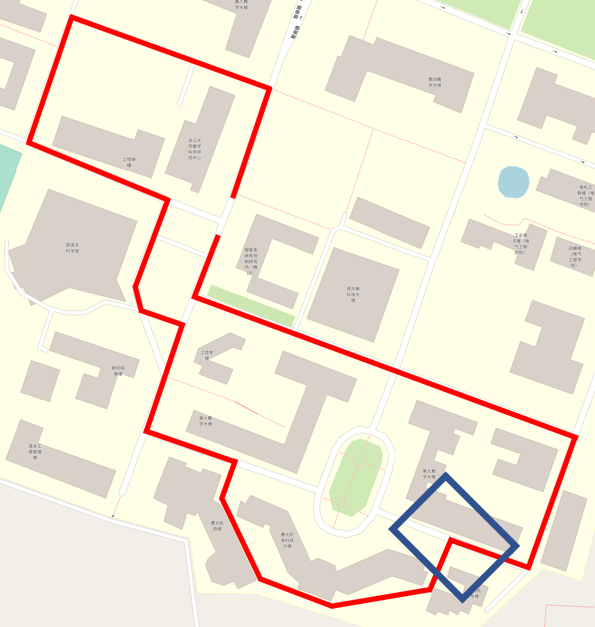}
\label{offline-map}
\end{minipage}%
}%
\subfigure[Online Cropped Routing Planning.]{
\begin{minipage}[htbp]{0.5\linewidth}
\centering
\includegraphics[width=0.5\textwidth]{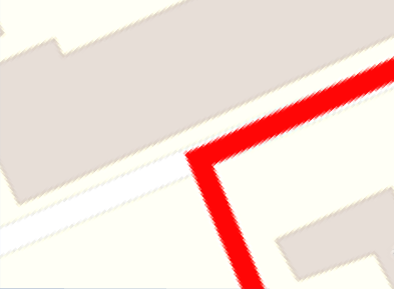}
\label{online-map}
\end{minipage}%
}%
\centering
\caption{Routing Planning Generation Schematic}
\end{figure}

In the driving intention module, GPS provides the current positioning and IMU provides the current yaw angle. Then, we crop an  online egocentric routing planning $R$ (as shown in Fig. A\ref{online-map}) according to the current position and yaw angle from a full offline routing planning (as shown in Fig. A\ref{offline-map}). The online egocentric routing planning $R$ is used as an input to the driving intention module. Note that in the driving intention module, we only need low cost GPS to get vehicle's location, which is explained in our previous work\cite{ma2020deepgoal}.

\section{Definition of Potential Maps}\label{apd:potential-map}
The driving intention map $I$ is generated by generator $\mathcal{G}$ whose final activation function is hyperbolic tangent function (tanh), so the value range of map $I$ is $(-1\sim1)$. For visualization, we map this driving intention $I$ to a grayscale map, where the white area represents the generated path (or called driving intention).

The potential map $C$ represents an area 25 meters wide by 50 meters long and the semantic information within this area around the vehicle in the top view. The vehicle is at the bottom center of the potential map and facing upwards towards the potential map. It is a $200\times400$ matrix or a grayscale map with each element or pixel having values from $0$ to $255$, representing whether the corresponding rasterized area ($0.125m\times0.125m$) is an obstacle, open space or driving intention. If an obstacle is detected by LiDAR in an area, the corresponding pixel is $0$. If the path obtained by driving intention module is projected to the area, the corresponding pixel is $255$. Otherwise the area is open space and the corresponding pixel is $127$.

\section{Controller Design}\label{apd:ctrl}
We use a feedback controller for throttle and brake control, and a nonlinear rear wheel feedback controller \cite{paden2016survey} for steering control:
\begin{equation}
\begin{aligned}
p_c(t) &= k_d e_d(t) + k_v e_v(t) + a_r(t) \\ 
\delta_c(t) &= \tan^{-1}\frac{\omega_c(t)L}{v(t)}
\end{aligned}
\end{equation}
where $L$ denotes the wheelbase, and at time $t$, $p_c$ represents throttle or brake (for $p_c>0$ throttle equals to $p_c$ and brake is $0$, for $p_c<0$ brake equals to $-p_c$ and throttle is $0$), $\delta_c$ is the steering angle, $\omega_c$ is the angular velocity,  $e_v$ is the tracking error between the current velocity $v$ and the reference velocity $v_r$, $e_d$ is the longitudinal error as shown in Fig.~\ref{ctrl}, $k_d$ and $k_v$ are controller gains for proper error feedbacks. Note that quantities with subscript $r$ are obtained from trajectory with time $t$. The angular velocity is calculated by:
\begin{equation}
\begin{aligned}
\omega_c(t) &= \frac{v_r(t)\kappa_r(t) \cos\theta_e(t)}{1-\kappa_r(t) e(t)}-(k_\theta|v_r(t)|)\theta_e(t)\\
&-(k_e v_r(t)\frac{\sin\theta_e(t)}{\theta_e(t)})e(t)
\end{aligned}
\end{equation}
where $e$ and $\theta_e$ are the lateral error and heading error as shown in Fig.~\ref{ctrl}, $\kappa_r$ is the curvature of the reference trajectory, $k_\theta$ and $k_e$ are the controller gains for proper error feedbacks.

\section{Prediction results on RoBotCar Dataset}\label{apd:robotcar-result}
\begin{figure}[h]
\centering
\subfigure[straight]{
\begin{minipage}[h]{0.32\linewidth}
\centering
\includegraphics[width=0.98\linewidth]{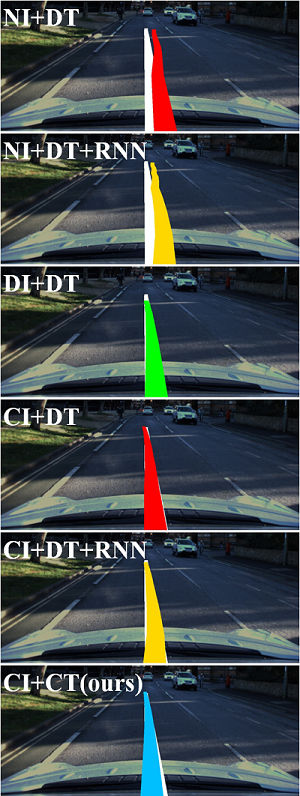}
\end{minipage}%
}%
\subfigure[turn]{
\begin{minipage}[h]{0.32\linewidth}
\centering
\includegraphics[width=0.98\linewidth]{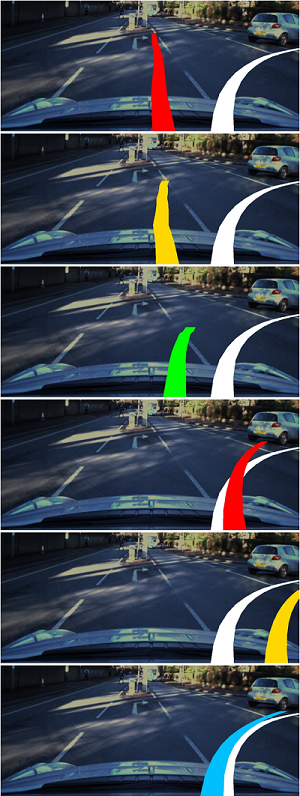}
\end{minipage}%
}%
\subfigure[stop]{
\begin{minipage}[h]{0.32\linewidth}
\centering
\includegraphics[width=0.98\linewidth]{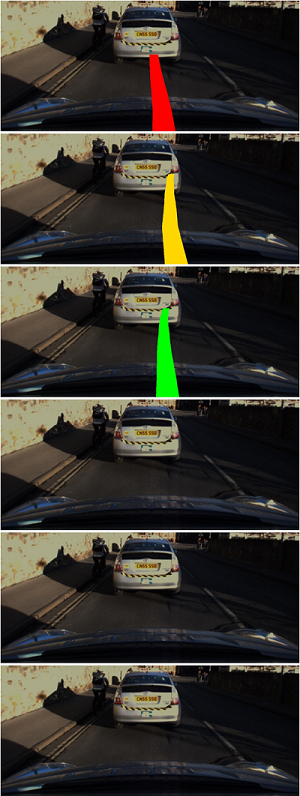}
\end{minipage}
}%
\centering
\caption{Prediction results on RoBotCar Dataset. {\textit{White}} trajectories denote ground truth and others correspond to predictions of comparison models.}
\label{oxford_result}
\vspace{-0.3cm}
\end{figure}

Fig \ref{oxford_result} displays typical qualitative results on these models. For various kinds of scenarios, our model generates the most human-like trajectory. Note figures in Fig \ref{oxford_result}(c) show no ground truth as the vehicle stops actually because of the front obstacle, and the performance of models is enhanced contributed from the use of semantic potential maps.

\section{Generalization Ability Experiment}\label{apd:generalization}
We used another dataset YQ21\footnote{\url{https://tangli.site/projects/academic/yq21/}} for testing generalization ability of different models.

\begin{figure}[htbp]
\centering
\subfigure[KITTI]{
\begin{minipage}[t]{0.3\linewidth}
\centering
\includegraphics[width=\textwidth]{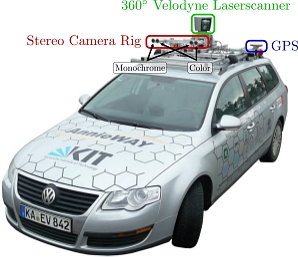}
\end{minipage}%
}%
\subfigure[RobotCar]{
\begin{minipage}[htbp]{0.3\linewidth}
\centering
\includegraphics[width=\textwidth]{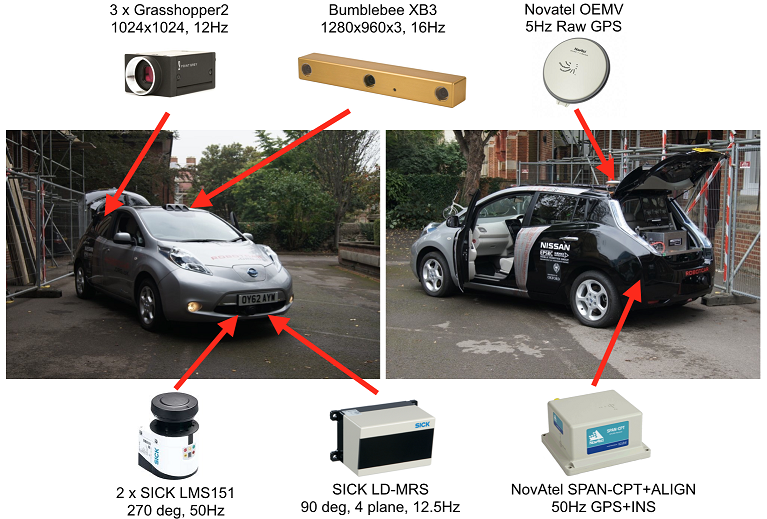}
\end{minipage}%
}%
\subfigure[YQ21]{
\begin{minipage}[htbp]{0.3\linewidth}
\centering
\includegraphics[width=\textwidth]{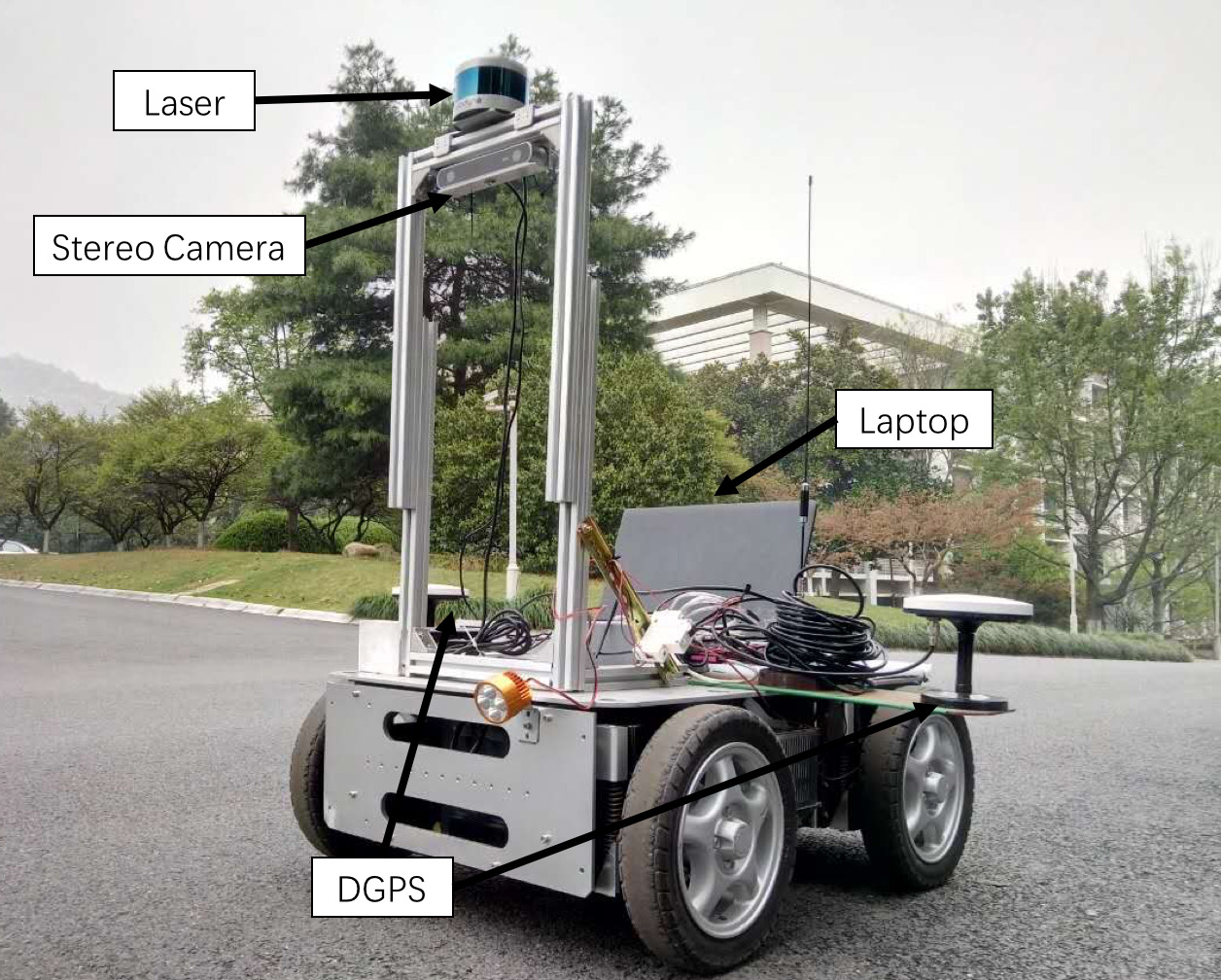}
\label{yq21vehicle}
\end{minipage}
}%
\centering
\caption{Devices Used for Different Datasets}
\end{figure}

In the YQ21 dataset, the robot shown in Fig. A\ref{yq21vehicle} was driven manually on the same route over $1~km$ in our campus at different time to collect data. $21$ sessions of sensory data were collected in spring as training set, while $3$ in autumn and $1$ in winter as test set. Since 2 types of real vehicles have been used in the KITTI and RobotCar datasets, we directly test the models on YQ21, without re-training the models. Since the YQ21 dataset has a high similarity to KITTI dataset in many settings, we directly transfer models trained on the KITTI dataset to the YQ21 dataset for testing the generalization ability. We compared our method with VTGNet\cite{cai2020vtgnet}, which is discrete intention input and discrete trajectory output method (DI+DT) without any explicit intermediate representation, and the experimental results are shown in Tab.\ref{generalization}, our method has stronger generalization ability.

\begin{table}[htbp]
\center
\caption{Generalization Experiments on YQ21 Dataset}
\label{generalization}
\begin{tabular}{c@{\ }c@{\ }c@{\ }c@{\ }c@{\ }c@{\ }c}
\toprule
Algorithm & $\mathcal{E}_{ad}$ & $\mathcal{E}_{x}$ & $\mathcal{E}_{y}$ & $\mathcal{E}_{fd}$& $\mathcal{E}_{v}$& $\mathcal{S}$\\
& $(m)$ & $(m)$ & $(m)$ & $(m)$& $(m/s)$& $(m/s^3)$\\
\midrule
VTGNet\cite{cai2020vtgnet} & $1.89$ & $1.64$ & $0.47$ & $3.17$ & $1.47$ & $0.90$\\
$\mathbf{Ours}$ & $\mathbf{1.47}$ & $\mathbf{1.17}$ & $\mathbf{0.43}$ & $\mathbf{1.65}$ & $\mathbf{0.69}$& $\mathbf{0.49}$\\
\bottomrule
\end{tabular}
\vspace{-0.2cm}
\end{table}

\section{Real-world Experiment}\label{apd:real-world}
The computation time of our method on different computational performance devices is shown in Tab.\ref{tab:time}. Our model has a shorter run time than VTGNet on the same device, even though our parameters are little larger. With a high performance GPU, the time used by our method can reach about $4 ms$, so it can directly achieve high frequency control. When using only CPU calculation, our method takes about $500 ms$, and if we do synchronous control, we can only achieve $2 Hz$ control frequency, while using asynchronous planning and control method proposed in this paper, we can still achieve high control frequency.
\begin{table}[htbp]
	\center
	\caption{Running Time and Parameter Size of Our Method and VTGNet}
	\label{tab:time}
	\begin{tabular}{c@{\ \ }c@{\ \ }c}
		\toprule
		Module  & VTGNet\cite{cai2020vtgnet} & Ours\\
		\midrule
		Running Time (CPU Only)   & $1087ms$ & $\mathbf{485ms}$ \\
		Running Time (CPU+GPU) & $13.3ms$ & $\mathbf{3.7ms}$ \\
		Parameters & $\mathbf{161.5M}$ & $179.6M$ \\
		\bottomrule
	\end{tabular}
\end{table}

\begin{figure}[htbp]
\center
\includegraphics[width=0.45\textwidth]{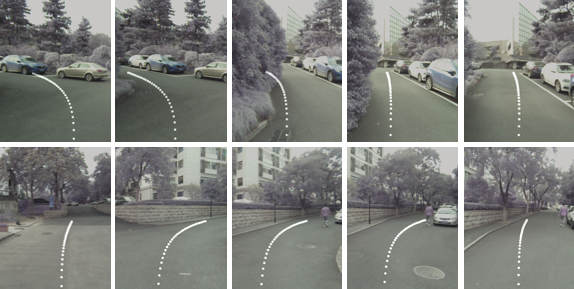}
\caption{Evaluation results of trajectory generation module transfer from RobotCar dataset to real world.}
\label{real-world-result}
\vspace{-0.3cm}
\end{figure}

\begin{figure}[htbp]
\centering
\subfigure[Human Interruptions Result of Our Method. 5 human interruptions occur totally.]{
\begin{minipage}[t]{0.45\linewidth}
\centering
\includegraphics[width=0.8\textwidth]{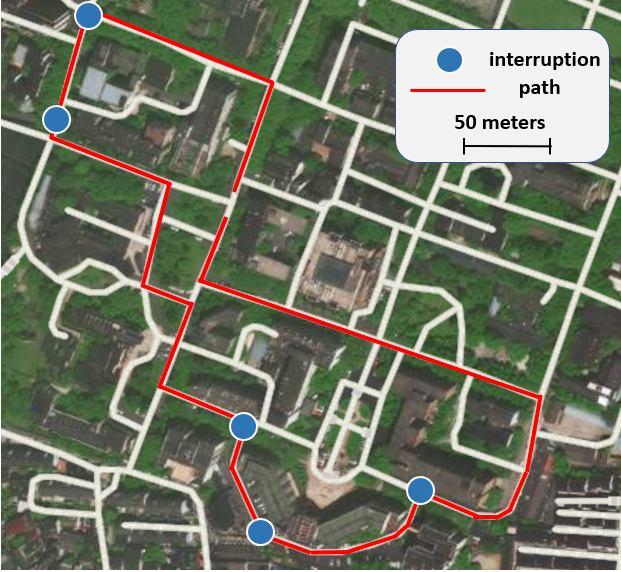}
\label{ours-real}
\end{minipage}%
}%
\subfigure[Human Interruptions Result of VTGNet. 8 human interruptions occur totally.]{
\begin{minipage}[t]{0.45\linewidth}
\centering
\includegraphics[width=0.8\textwidth]{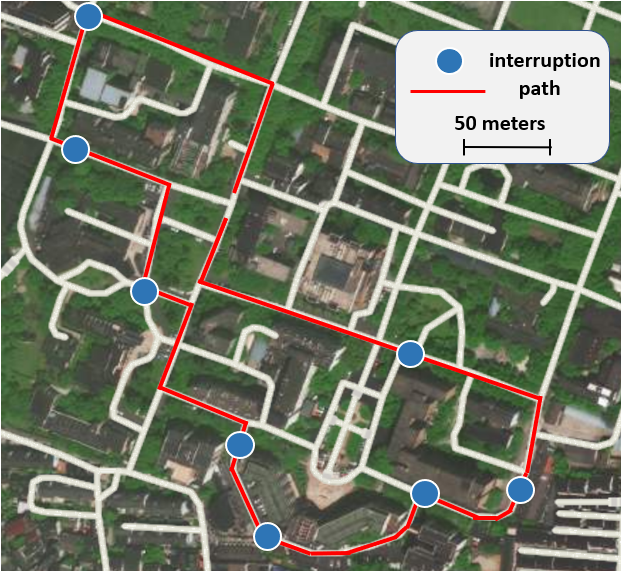}
\label{vtgnet-real}
\end{minipage}%
}%
\centering
\caption{Human Interruptions Result of Our Method and VTGNet. Most human interruptions occur at intersections. Our method has fewer human interruptions.}
\end{figure}

We also train our model and VTGNet with real-world data, and test the two models in real-world closed-loop experiment, counting the number of human interruptions. The result is shown in Tab. \ref{tab:real-world}, and the trajectories generated by our methods are shown in Fig.~\ref{real-world-result}. VTGNet has 8 human interruptions occur totally, while our method has 5 human interruptions occur totally, which is also shown in Fig. A\ref{ours-real} and Fig. A\ref{vtgnet-real}. Hence, our method also has long interruption distance and less driving time, which proves the superiority of our method.

\begin{table}[htbp]
	\center
	\caption{Real-world Statistics of Our Method and VTGNet}
	\label{tab:real-world}
	\begin{tabular}{c@{\ \ }c@{\ \ }c}
		\toprule
		Module  & VTGNet\cite{cai2020vtgnet} & Ours\\
		\midrule
		Number of Human Interruptions  & $8$ & $\mathbf{5}$ \\
		Average Distance Between Interruption & $138~m$ & $\mathbf{220~m}$ \\
		Driving Time & $783~s$ & $\mathbf{672~s}$ \\
		\bottomrule
	\end{tabular}
\end{table}

\section{Network Architecture}\label{apd:network}
The network structure is shown in Tab. ~\ref{generator}, Tab. ~\ref{discriminator}, and Tab. ~\ref{trajectory}.

\begin{table*}[htbp]
	\caption{Architecture of the Generator}
	\label{generator}
	\centering
	\resizebox{440pt}{!}{\begin{tabular}{ccccccccc}
		\toprule
		Layer & Channels In & Channels Out & Kernel & Stride & Padding & Normalize & dropout & Activation\\
		\midrule
		Conv1 & 6 & 64 & $4 \times 4$ & 2 & 1 & No & None & LeakyReLU \\
        Conv2 & 64  & 128 & $4 \times 4$ & 2 & 1 &Yes& None & LeakyReLU \\
        Conv3 & 128 & 256 & $4 \times 4$ & 2 & 1 & Yes & None & LeakyReLU \\
        Conv4 & 256 & 512 & $4 \times 4$ & 2 & 1 & Yes & 0.5 & LeakyReLU \\
        Conv5 & 512 & 512 & $4 \times 4$ & 2 & 1 & Yes & 0.5 & LeakyReLU \\
        Conv6 & 512 & 512 & $4 \times 4$ & 2 & 1 & Yes & 0.5 & LeakyReLU \\
        Conv7 & 512 & 512 & $4 \times 4$ & 2 & 1 & No & 0.5 & LeakyReLU \\
        UpConv1 & 512 & 512 & $4 \times 4$ & 2 & 1 & Yes & 0.5 & ReLU \\
        UpConv2 & 1024 & 512 & $4 \times 4$ & 2 & 1 & Yes & 0.5 & ReLU \\
        UpConv3 & 1024 & 512 & $4 \times 4$ & 2 & 1 & Yes & 0.5 & ReLU \\
        UpConv4 & 1024 & 256 & $4 \times 4$ & 2 & 1 & Yes & None & ReLU \\
        UpConv5 & 512 & 128 & $4 \times 4$ & 2 & 1 & Yes & None & ReLU \\
        UpConv6 & 256 & 64 & $4 \times 4$ & 2 & 1 & Yes & None & ReLU \\
        Conv8 & 128 & 1 & $4 \times 4$ & 1 & 1 & No & None & Tanh \\
		\bottomrule
	\end{tabular}}
\end{table*}

\begin{table*}[htbp]
	\caption{Architecture of the Discriminator}
	\label{discriminator}
	\centering
	\resizebox{440pt}{!}{\begin{tabular}{ccccccccc}
		\toprule
		Layer & Channels In & Channels Out & Kernel & Stride & Padding & Normalize & dropout & Activation\\
		\midrule
		Conv1 & 7 & 64 & $4 \times 4$ & 2 & 1 & No & None & LeakyReLU \\
        Conv2 & 64  & 128 & $4 \times 4$ & 2 & 1 &Yes& None & LeakyReLU \\
        Conv3 & 128 & 256 & $4 \times 4$ & 2 & 1 & Yes & None & LeakyReLU \\
        Conv4 & 256 & 512 & $4 \times 4$ & 2 & 1 & Yes & None & LeakyReLU \\
        Conv5 & 512 & 1 & $4 \times 4$ & 1 & 1 & No & None & None \\
		\bottomrule
	\end{tabular}}
\end{table*}

\begin{table*}[htbp]
	\caption{Architecture of the Trajectory Generation Model}
	\label{trajectory}
	\centering
	\resizebox{440pt}{!}{\begin{tabular}{ccccccccc}
		\toprule
		Layer & Channels In & Channels Out & Kernel & Stride & Padding & Normalize & dropout & Activation\\
		\midrule
		Conv1 & 1 & 64 & $5 \times 5$ & 3 & 2 & No & None & LeakyReLU \\
		Max Pool & 64 & 64 & None & None & None & No & None & None \\
        Conv2 & 64  & 128 & $5 \times 5$ & 4 & 2 &Yes& None & LeakyReLU \\
        Max Pool & 128 & 128 & None & None & None & No & None & None \\
        Conv3 & 128 & 256 & $3 \times 3$ & 2 & 1 & Yes & None & LeakyReLU \\
        Max Pool & 256 & 256 & None & None & None & No & None & None \\
        Conv4 & 256 & 256 & $3 \times 3$ & 2 & 1 & Yes & None & None \\
        GRU & 256 & 256 & None & None & None & No & 0.2 & LeakyReLU \\
        FC1 & 258 & 512 & None & None & None & No & None & Tanh \\
        FC2 & 512 & 512 & None & None & None & No & None & Tanh \\
        FC3 & 512 & 512 & None & None & None & No & None & Tanh \\
        FC4 & 512 & 256 & None & None & None & No & None & Cos \\
        FC5 & 256 & 2 & None & None & None & No & None & None \\
		\bottomrule
	\end{tabular}}
\end{table*}

\end{appendices}
\end{document}